\documentclass[review]{elsarticle}
\usepackage{booktabs,caption,threeparttable}
\usepackage{adjustbox}
\usepackage{lineno,hyperref}
\usepackage{amsmath,amssymb,amsfonts}
\journal{Pattern Recognition}
\usepackage{algorithmic}
\usepackage{multirow}
\usepackage{array}
\usepackage{xcolor}
\usepackage{soul}
\usepackage{arydshln}
\usepackage[demo]{rotating}
\usepackage{ulem}
\usepackage{kotex}
\usepackage{mathtools}
\colorlet{tdcolor}{blue!8}
\sethlcolor{tdcolor}
\usepackage{bigstrut}

\DeclarePairedDelimiter{\abs}{\lvert}{\rvert}

\begin{document}

\begin{frontmatter}

\title{Spatial Reasoning for Few-Shot Object Detection}

\author[mymainaddress]{Geonuk Kim\fnref{equal}}
\author[mymainaddress]{Hong-Gyu Jung\fnref{equal}}
\author[mymainaddress,mysecondaryaddress]{Seong-Whan Lee\corref{mycorrespondingauthor}}
\fntext[equal]{Equal contribution}
\cortext[mycorrespondingauthor]{Corresponding author}
\ead{sw.lee@korea.ac.kr}

\address[mymainaddress]{Department of Brain and Cognitive Engineering, Korea University, Anam-dong, Seongbuk-gu, Seoul, 02841, Korea}
\address[mysecondaryaddress]{Department of Artificial Intelligence, Korea University, Anam-dong, \\ Seongbuk-gu, Seoul, 02841, Korea}

\begin{abstract}
Although modern object detectors rely heavily on a significant amount of training data, humans can easily detect novel objects using a few training examples. The mechanism of the human visual system is to interpret spatial relationships among various objects and this process enables us to exploit contextual information by considering the co-occurrence of objects. Thus, we propose a spatial reasoning framework that detects novel objects with only a few training examples in a context. We infer geometric relatedness between novel and base RoIs (Region-of-Interests) to enhance the feature representation of novel categories using an object detector well trained on base categories. We employ a graph convolutional network as the RoIs and their relatedness are defined as nodes and edges, respectively. Furthermore, we present spatial data augmentation to overcome the few-shot environment where all objects and bounding boxes in an image are resized randomly. Using the PASCAL VOC and MS COCO datasets, we demonstrate that the proposed method significantly outperforms the state-of-the-art methods and verify its efficacy through extensive ablation studies.
\end{abstract}


\end{frontmatter}


\section{Introduction}
Learning to classify and localize each object in an image is a fundamental research problem among various recognition areas \cite{yang2007reconstruction, he2016deep, suk2008recognizing, long2015fully, lee1999integrated}. Owing to considerable advances in deep learning, the object detection problem has been addressed successfully by milestone works such as Faster-RCNN \cite{ren2015faster}, Mask-RCNN \cite{he2017mask}, YOLO \cite{redmon2016you} and SSD \cite{liu2016ssd}. However, modern object detectors \cite{WANG2021107593,XU2020107098} rely significantly on a large amount of training data including category labels and bounding boxes for each object. Collecting such large-scale training data incurs high costs and indicates that a recognition system should be re-trained with considerable complexity when adding novel object categories.

To address this problem, few-shot learning \cite{snell2017prototypical,  finn2017model, jung2020few} has recently gained significant attention, and several approaches \cite{kang2019few,wang2019meta,yan2019meta,wang2020frustratingly,xiao2020few,wu2020multi} have been applied to develop \textit{few-shot object detectors} for detecting data-scarce novel categories and data-sufficient base categories. For example, pioneering works \cite{kang2019few,wang2019meta,yan2019meta,xiao2020few} proposed to attach a meta-learner to an existing object detector trained on base categories. The meta-learner generates category attention vectors that are used to re-weight the feature maps with a few training examples. This results in category-discriminative feature maps that remodel a prediction layer to deal with both novel and base categories. On the other hand, a recent work \cite{wang2020frustratingly} proposed a two-stage approach where the first stage trains base categories with large-scale training data and the second stage fine-tunes only the classification layer to further detect novel categories with a few training examples. A data augmentation approach \cite{wu2020multi} was presented to address limited scale variations among a few training examples.

However, the existing methods mainly focus on manipulating an object-level feature, and thus the detection performance solely depends on the discriminatory capability of Region-of-Interest (RoI) features. On the other hand, it is well-known that humans recognize objects in a context by interpreting the relationships among objects \cite{oliva2007role,auckland2007nontarget,lee1990translation}. 
 \begin{figure}
\begin{center}
\includegraphics[width=1.0\linewidth,height=4.5cm]{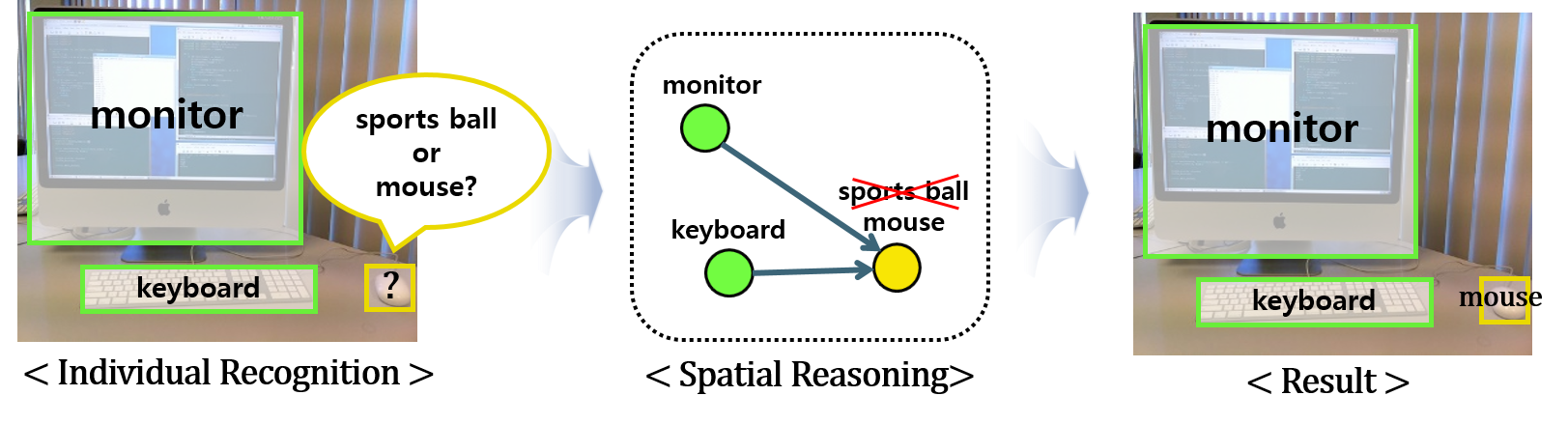}
\end{center}
\caption{{Conceptual example of the human visual system for recognizing new objects using a spatial reasoning process. Even if a human has little experience of new objects and has trouble recognizing the objects individually, it is possible to recognize the objects applying a spatial reasoning process. Specifically, human re-interpret ambiguous objects by considering other objects and the geometric patterns among them.}}\label{fig:human_vision}
\end{figure}
For example, as shown in Fig. \ref{fig:human_vision}, humans try to recognize a mouse on a desk with the following reasoning process: ``It looks like a sports ball or a mouse, and the object is near the monitor and keyboard. Thus, it is more likely to be a mouse than a sports ball''. 

Inspired by the human visual system, we introduce a few-shot object detector with a spatial reasoning framework (FSOD-SR). In this scheme, a novel object is predicted in a context rather than merely using a local RoI feature. We take into account the co-occurrence of objects using spatial information with base categories to improve the feature representation of novel categories. Consideration of the inter-relationships of objects is particularly useful for few-shot learning as a model can explicitly learn where to pay attention to several objects in an image.

We define a spatial graph where nodes and edges represent individual RoIs and their relatedness, respectively. Here, the nodes are simply obtained by a region proposal network (RPN) and belong to a base or novel category. Then, the edges are calculated based on the box coordinates of each RoI and the similarity scores that indicate how an object is related to specific base categories. Given the components of a graph, we employ a graph convolutional network (GCN) to enhance the discriminative power of RoIs by propagating the information contained in each node. Finally, the aggregated features from the GCN are concatenated with the original RoI features to reflect the inter-relationships among objects.

Meanwhile, we propose spatial data augmentation to enlarge the training examples for the aforementioned process. In this strategy, each object region is resized $T$-times simultaneously to augment RoI samples and this helps to efficiently find the object relatedness using the various sizes of objects. Furthermore, this object-wise strategy enlarges training examples exponentially as a multi-set permutation problem.

Our contributions are summarized as follows:
\begin{enumerate}
    \item We propose a spatial reasoning process to detect novel categories in a context that is less considered in few-shot object detection. 
    \item We propose a spatial data augmentation method to overcome few-shot environments. Our approach augments RoI samples to optimize the spatial reasoning process with various sizes of objects.
    \item We conduct extensive experiments on the PASCAL VOC and MS COCO datasets, which are widely used for object detection and show that the proposed method significantly outperforms state-of-the-art methods.
\end{enumerate}

\section{Related work}

\subsection{Object detection}
Object detection is a core problem in computer vision and significant progress has been made in recent years using convolutional neural networks (CNNs). Modern object detectors are divided into two approaches: a two-stage and one-stage object detectors. A two-stage object detector uses an RPN to generate RoIs in the first stage and clarifies the RoIs with a classifier and a bounding box regression in the second stage. The R-CNN series \cite{ren2015faster,he2017mask} represents a two-stage object detector. One-stage object detectors, including YOLO \cite{redmon2016you}, SSD \cite{liu2016ssd}, and their variants, directly classify and localize an object without the RPN, thus requiring lower computation complexity than the two-stage detectors. However, the aforementioned methods have the disadvantage of requiring a large-scale training dataset to be used in practical applications.

\subsection{Few-shot object detection}
Few-shot learning aims to recognize novel categories utilizing a few training examples. To solve this problem, most researchers have focused on few-shot classification where an object is clearly placed in an image \cite{zhu112temperature,huang2020behavior, HUANG2021107935}. {In other words, the approaches do not consider the concept of backgrounds and bounding box regression. On the other hand, few-shot object detectors where the goal is to detect multiple objects of various sizes and locations have been recently developed to address a more challenging problem.} 

{For example, FSRW }\cite{kang2019few}{ uses a sub-network to predict categorical attention vectors that are exploited to re-wight feature maps from YOLOv2 }\cite{redmon2017yolo9000}{. Then, a prediction layer is re-modeled based on the category-attentive feature maps. To ensure the attention vector is generalizable to unseen categories, a meta learning approach is adopted during training base categories. As a result, this process enables a model to learn where to attention with only a few training examples for each category using a re-weighting network. At the same time, Meta R-CNN }\cite{yan2019meta}{ and FSIW }\cite{xiao2020few}{ were developed with the same motivation but use Faster R-CNN }\cite{ren2015faster}{ as a backbone. Instead of using an attention mechanism, MetaDet }\cite{wang2019meta}{ aims at directly generating classification and regression weights given an image. This approach utilizes information from base categories with a large-scale dataset to produce the weights of novel categories. MPSR }\cite{wu2020multi}{ tackled an improper scale issue where positive RoI samples in a test image (e.g., the face of a dog) can be treated as negetive RoI samples in a training image (e.g., the whole body of a dog) and argued that it becomes worse in few-shot environments. To suppress the effect of improper negative samples, an object-wise multi-scale feature extractor was developed to increase the number of positive samples using an image pyramid.}

{Overall, the previous approaches train an auxiliary branch on base categories, apply it to the output of a backbone, and fine-tune object detectors to adapt to novel categories. Meanwhile, TFA }\cite{wang2020frustratingly}{ revealed that simply adding classification and regression weights to the last layer of existing detectors and only fine-tuning the weights significantly improved the few-shot detection performance without any auxiliary branch.}

{However, the existing methods focus on how to extract discriminative features in terms of a single RoI and }do not consider the innate properties present in an image. In general, the PASCAL VOC and MS COCO datasets are composed of real-life images where multiple objects co-exist. Thus, there is an opportunity to further exploit the relationships among objects as additional information to overcome a few-shot environment. To address this issue, we propose a spatial reasoning framework to consider the co-occurrence property of objects.

\subsection{Visual reasoning} 
Reasoning the relationship among objects has been studied for various tasks such as image classification and object detection to explore contextual information. Pioneering studies \cite{reed2016learning,lampert2009learning} relied on linguistic information using a natural language corpus rather than explicitly considering various features within images. Recent studies have been conducted to reason visual relationships directly from images. In \cite{liu2018structure}, the object detection problem was defined as node detection in a graph, based on the fact that objects in an image exist in a specific contextual structure. To this end, a dense graph was trained in which all objects were connected using several gated recurrent unit cells. RelationNet \cite{hu2018relation} also exploits a dense graph but uses an attention mechanism to learn where to focus in an image. However, the dense graph approaches, where all objects are connected, caused redundant message transfers among graph nodes. To solve this issue, SGRN \cite{xu2019spatial} and Graph-RCNN \cite{yang2018graph} tried to represent a sparse graph by sampling only the top-$k$ maximum values from the edge candidates. However, as the number of objects varies for each image, there is a limitation by selecting a pre-defined number of edges not to consider clearly the characteristics of each image. Thus, we propose to transform redundant edges into negative values in a latent space to remove the edges dynamically through the ReLU function. Furthermore, as previous approaches require a large-scale dataset to model spatial relationships among objects, spatial reasoning for novel categories with only a few training examples cannot be guaranteed as the detectors are trained to be generalized to \textit{unseen images} but not \textit{unseen categories}. To address the issue, we measure how novel categories are related to base categories in a classification distribution and propose a graph neural network to exploit the relationships. Also, we present a spatial data augmentation technique where each object in an image is randomly resized to overcome few-shot environments.

 \begin{sidewaysfigure}
\begin{center}
\includegraphics[height=5.5cm]{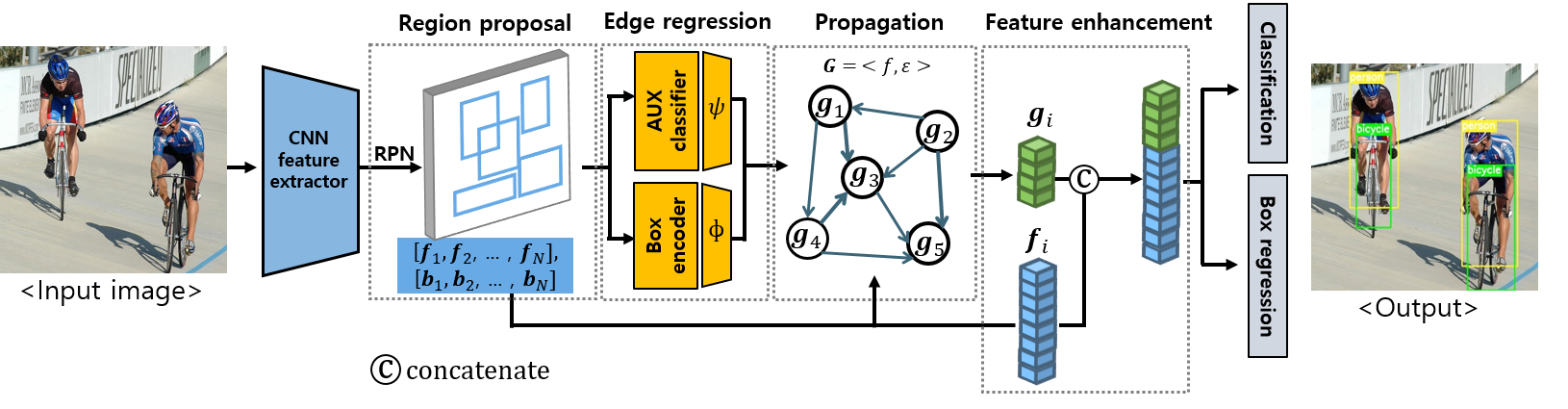}
\end{center}
\caption{An Overview of the proposed FSOD-SR. We exploit a graphical representation of the relatedness among RoIs to detect novel objects in a context. The spatial relation is formulated as a region-to-region graph where nodes $f$ and edges $\varepsilon$ represent RoIs and the relatedness between RoIs. The edges are constructed based on box coordinates and the degree of similarity with base categories. Given the graph, we employ a graph convolutional network to enhance the discriminative power of RoIs.} \label{fig:overall}
\end{sidewaysfigure}

\section{Proposed method}
In this study, we consider two training phases. In the base training phase, we train Faster-RCNN \cite{ren2015faster} on base categories $C_{base}$ with a large-scale dataset $D_{base}$. In the fine-tuning phase, we aim to additionally detect novel categories $C_{novel}$ with only a few training examples $D_{novel}$. To this end, we fine-tune the detector with $k$ bounding box annotations for each object sampled from $D_{base}$ and $D_{novel}$ for few-shot learning. It is worth noting that the $C_{base}$ and $C_{novel}$ categories are disjoint. Finally, in the test phase, we report the detection performances on $D_{test}$ that is composed of base and novel categories. In the following sections, we elaborate on the proposed spatial reasoning and data augmentation methods.

\subsection{Overall process}
An overview of the proposed FSOD-SR is depicted in Fig. \ref{fig:overall}. We develop a few-shot object detection method that can be implemented on any region based object detector such as Faster-RCNN \cite{ren2015faster}. In our method, the spatial relation is formulated as a region-to-region graph $G: G=<f,\varepsilon>$. First, projection functions $\psi$ and $\phi$ dynamically learn a sparse adjacency matrix from the RoI features and the corresponding coordinates. Using the constructed graph $G$, we employ graph convolution to propagate RoI features into neighborhoods of each region. Then, an aggregated feature $g$ is concatenated to the original RoI feature $f$ to improve the feature representation of the region. To further enrich the training examples, we augment the training data from $k$-shot examples of the $D_{base}$ and $D_{novel}$ in terms of objects in an image.

\subsection{Spatial reasoning} \label{sect:SR}
Spatial reasoning aims to exploit a graphical representation of the relatedness among RoIs to enhance the feature representation of each region, which is particularly useful for ambiguous novel RoIs. We formulate the spatial relation as a region-to-region graph $G: G=<f,\varepsilon>$ where nodes $f$ and edges $\varepsilon$ represent RoIs and the relatedness between RoIs, respectively. Given the $N$ number of $d$ dimensional RoI visual features $f\in \mathbb{R}^{N \times d}$ and the corresponding coordinates $b\in \mathbb{R}^{N \times 4}$ from an RPN, we first transform $f_{i}$ for each region $i$ into the classification distribution $c_i\in \mathbb{R}^{1 \times (1+C_{base})}$ using an auxiliary classifier for base categories. In addition, a box encoder transforms $b_i=(x_{min},y_{min},x_{max},y_{max})$ into a 6-dimensional vector $o_i$ to normalize the size of an RoI as follows:
\begin{equation} 
o_i=\left[\frac{x_{min}}{w_{img}}, \frac{y_{min}}{h_{img}}, \frac{x_{max}}{w_{img}},\frac{y_{max}}{h_{img}}, \frac{(x_{max}-x_{min})\cdot(y_{max}-y_{min})}{w_{img} \cdot h_{img}}, \frac{y_{max}-y_{min}}{x_{max}-x_{min}}\right], \label{eqn:box_encoding}
\end{equation}
where $w_{img}$ and $h_{img}$ denote the width and height of an image, respectively.

\begin{figure}
    \begin{center}
    \includegraphics[width=1.0\linewidth,height=5.1cm]{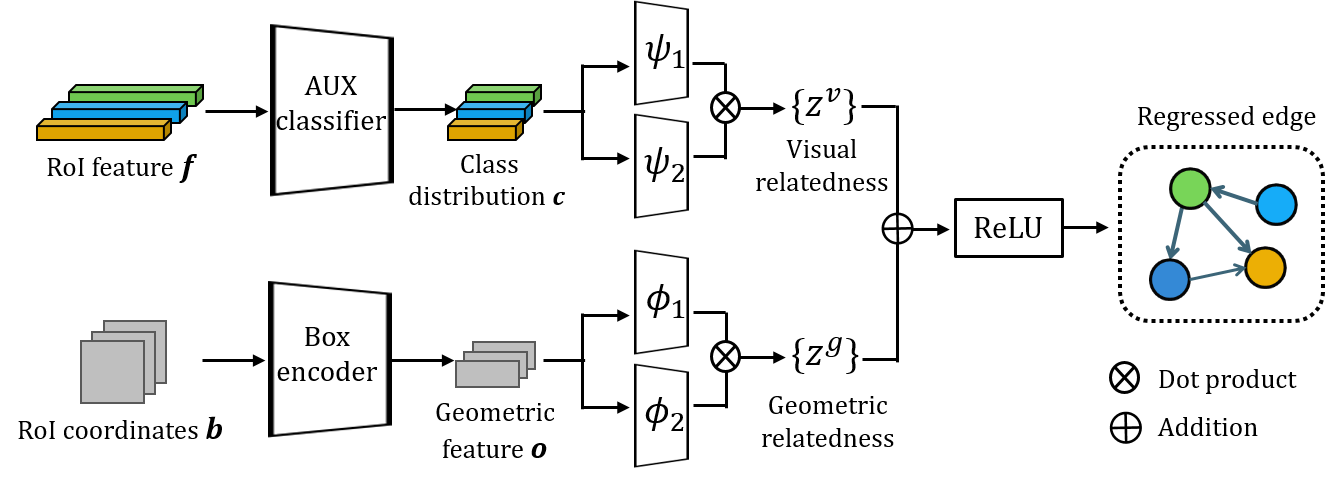}
    \end{center}
    \caption{Detailed flowchart of the edge regression. We transform the RoI visual features and the corresponding coordinates from an RPN into latent spaces to calculate the relatedness of RoIs. Finally, the ReLU function is exploited to build a sparse graph.} \label{fig:edge_reg}
\end{figure}

To represent the visual and geometric relatedness between each RoI pair, we project the class distribution $\mathbf{c}$ and the geometric feature $\textbf{o}$ to the latent spaces $z^{v}\in \mathbb{R}^{N \times N}$ and $z^{g}\in \mathbb{R}^{N \times N}$ as follows:
\begin{equation} 
\begin{aligned}
z_{i,j}^{v} = {\psi_1(c_i)} \cdot {\psi_2(c_j)}^T \\
z_{i,j}^{g} = \phi_1(o_i) \cdot {\phi_2(o_j)}^T,
\end{aligned}\label{eqn:latent_construct}
\end{equation}
where $\psi$ and $\phi$ denote the projection functions whose weight matrix represents $W_{\psi}\in \mathbb{R}^{(1+C_{base}) \times p}$ and $W_{\phi}\in \mathbb{R}^{6 \times q}$, respectively. The projection functions are separated (e.g., $\psi_1,\psi_2$) by an identical structure to represent an asymmetric form. Then, the final edge ${\varepsilon} \in \mathbb{R}^{N \times N}$ is computed by considering both $z^v$ and $z^g$ as follows:
\begin{equation} 
\varepsilon_{i,j} = \sigma\left(z_{i,j}^{v} + z_{i,j}^{g} \right),\label{eqn:edge_regression}
\end{equation}
where we apply the ReLU function $\sigma$ to prune out noisy edges that represent redundant connections. Note that this zero-trimming operation dynamically helps to build a sparse adjacency matrix $\varepsilon$. The flowchart of the edge regression is illustrated in Fig. \ref{fig:edge_reg}.

Given a graph $G=<f,\varepsilon>$, we exploit the GCN to aggregate the contextual information of each region $i$ from its neighborhood. The aggregation is covered by $L$-layer graph convolutional blocks and each node is initialized with an RoI feature $f$ as follows:
\begin{equation}
H_i^{0}=f_i
\label{eqn:node_init}
\end{equation}

At each layer $l$, each node $i$ aggregates features from its neighborhoods, and this is expressed as
\begin{equation}
H_{i}^{l} = \sigma \left(H_{i}^{l-1}+\sum_{j\in Neighbor(i)}^{} {D_{i,i}^{-1}\varepsilon}_{i,j}H_{j}^{l-1}\Theta^{l-1} \right),\label{eqn:gcn}
\end{equation}
where $H^{l}$ and $\Theta^{l}$ indicate the feature representation for the nodes and a trainable weight matrix in the $l$-th layer, respectively. $D\in \mathbb{R}^{N \times N}$ is a degree matrix that normalizes each row in $\varepsilon$ to ensure that the scale of the feature representations is not modified by aggregation \cite{kipf2016semi}.

Then, we define aggregated features $g\in \mathbb{R}^{N \times q}$ as the $q$-dimensional feature representation for the nodes at the last layer, $l=L$,
\begin{equation}
g_i=H_i^{L}.
\label{eqn:node_end}
\end{equation}

Finally, an aggregated feature $g_i$ is concatenated with $f_i$ to enhance the feature representation of a region $i$. Then, we use a classification score based on the cosine similarity to classify the concatenated feature as follows:
\begin{equation}
P_{i,u} =  \alpha \frac{{e_i^T}\cdot{w_u}}{{\abs*{e_i}\cdot{\abs*{w_u}}}},\label{eqn:cos_cls}
\end{equation}
where $e_i = f_i||g_i$ and `$||$' denotes the concatenation operation. ${P_{i,u}}$ is the classification score between the enhanced feature $e_i$ and the weight vector $w_u$ of a category $u$. $\alpha$ is a scaling factor for stability.

\subsection{Spatial data augmentation}
\begin{figure*}
\begin{center}
\includegraphics[width=1.0\linewidth,height=5cm]{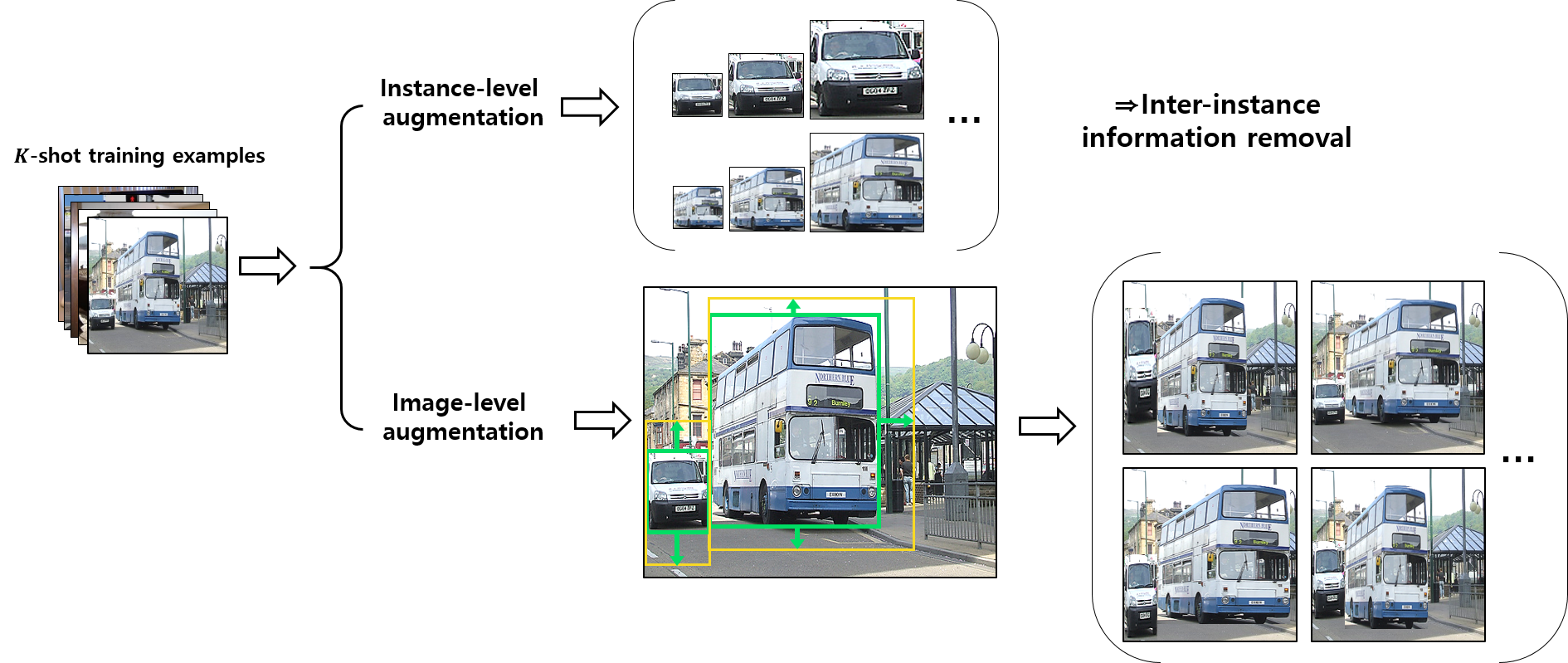}\end{center}
\caption{Example of the proposed image-level spatial augmentation and the existing instance-level augmentation \cite{wu2020multi}. As the instance-level augmentation uses the augmented instances as an input of an object detector, it cannot fully exploit the inter-relationships among objects (e.g., ``bus'' next to ``car''). Thus, we propose an image-level augmentation as we randomly resize the objects in an image and use the entire images as an input. In addition, unlike the instance-level augmentation, our method exponentially enlarges the training examples according to the number of objects in an image and their augmented sizes.}\label{fig:spatial_aug}
\end{figure*}

To enrich training examples for $C_{novel}$, one possible solution is to extract objects independently, resize them to various scales and feed only the objects to an object detector \cite{wu2020multi}. However, such an instance-level augmentation removes co-occurrence information (e.g., ``bus'' next to ``car'') because only the resized RoI patches are directly fed to the prediction layer. Thus, to fully exploit the information in the $k$-shot images, we propose to augment training examples at the image level. As shown in Fig. \ref{fig:spatial_aug}, we resize each region randomly $T$-times among the regions simultaneously and attach them to the original image. Thus, when $\{{x'_n}\}_{n=1}^{M}$ represents a set of augmented images from a single image $x$, the number of augmented images $M$ can be formulated as
\begin{equation}
    M={T^S},\label{eqn:aug_form}
\end{equation}
where $S$ denotes the number of objects in an image $x$. As shown in Eq. \ref{eqn:aug_form}, $M$ increases exponentially with the number of objects. We exploit both augmented and original images to train our model.

\subsection{Training strategy} \label{sect:TS}
\subsubsection{Base training}
We first train Faster-RCNN \cite{ren2015faster} with the spatial reasoning process to detect base objects $C_{base}$. This phase allows the detector to prepare how to regress edges between RoIs and propagate the relevant features. To optimize the proposed model, we use the cross entropy loss and the smooth $L1$ loss $\hat{L}_1$ for the classification and bounding box regression layers, respectively. The loss function of the detection head is formulated as
\begin{equation}
    Loss=-\frac{1}{N}\sum_{i=1}^{N} y_i^*\left[logP(y_i|e_i)+log{P}(y^{aux}_i|f_i)\right]+\\
    \frac{1}{N}\sum_{i=1}^{N}{\hat{L}_1}\left(B(e_i)-t_i^*\right),\label{eqn:loss_base}
\end{equation}
where $y$ and $y^*$ denote the predicted and ground-truth categories, respectively. $B(\cdot)$ and $t^*$ indicate the regressed box coordinates and the ground-truth box coordinates, respectively. Finally, we use an auxiliary classifier to produce $y^{aux}$ for $\mathbf{c}$ in Eq. \ref{eqn:latent_construct}. The RPN is trained as in Faster-RCNN \cite{ren2015faster}. 

\subsubsection{Few-shot fine-tuning}
After the base training phase, only the classification layer of the detection head is replaced to classify both $C_{base}$ and $C_{novel}$ with $k$-shot training examples. In this phase, augmented images are considered by randomly resizing each region three times. Thus, the loss function is formulated as
\begin{equation}
\begin{aligned}
    Loss=-\frac{1}{N}\sum_{i=1}^{N} y_i^*\left[{logP(y_i|e_i)+log{P}(y_i|e'_i))}\right]+\\
    \frac{1}{N}\sum_{i=1}^{N}\left[\hat{L}_1\left(B(e_i)-t_i^*\right)+\hat{L}_1\left({B(e'_i)}-{t'}_i^*\right)\right],
\end{aligned}\label{eqn:loss_ft}
\end{equation} 
where ${t'}_i^*$ indicates the ground-truth box coordinates from the augmented training images. Here, $e'_i$ represents $f'_i||g'_i$ where $f'_i$ and $g'_i$ are the features from the augmented training images.

\begin{figure}[t]
    \centering
    \begin{center}
    \includegraphics[width=1.0\linewidth,height=10cm]{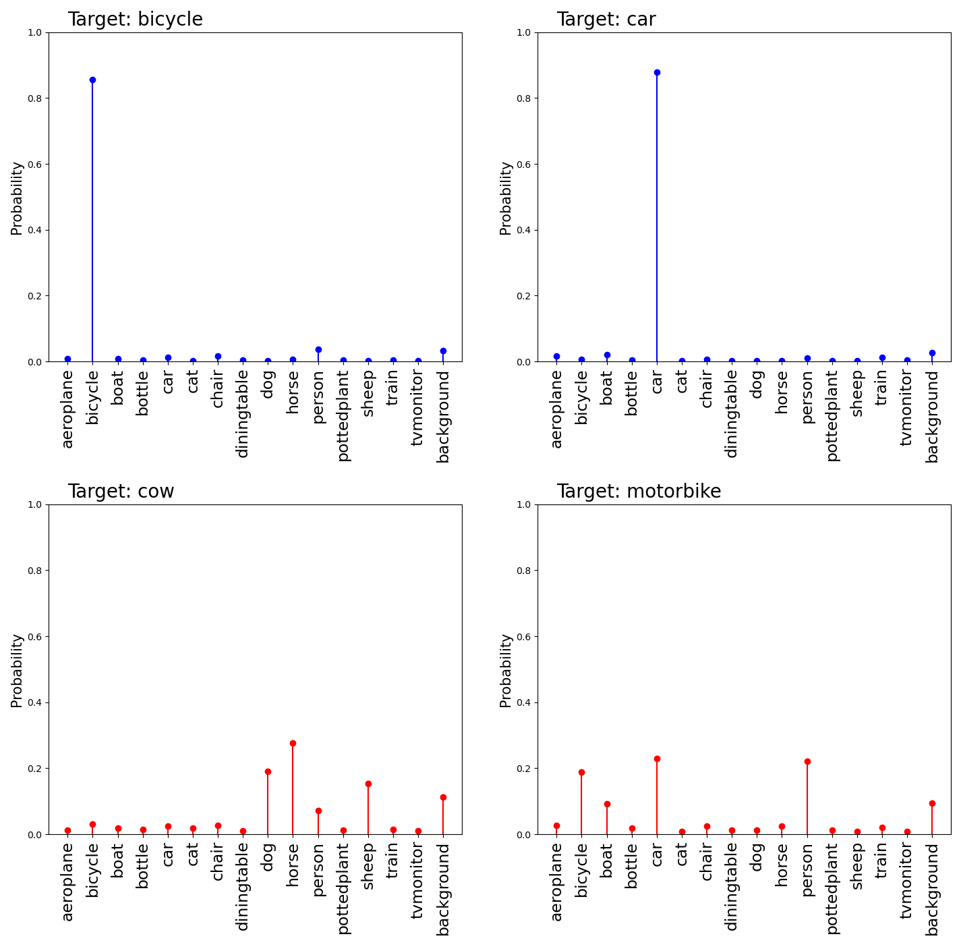}
    \end{center}
    \caption{Classification distribution for base and novel categories using an object detector trained on base categories. The values are averaged using the PASCAL VOC 2007 test set. The top figures represent the classification results using base categories. The bottom figures are the results of novel categories. Based on the unseen ``cow'' and ``motorbike'' categories, it tends to classify them as similar base categories rather than backgrounds.} \label{fig:bs2nv}
\end{figure}

\subsection{Discussion} \label{sect:discuss}
The difference between the base training and the few-shot fine-tuning phases is whether to (1) train the auxiliary classifier and (2) use the augmented images. In other words, we still exploit the classification distribution $c_i\in \mathbb{R}^{1 \times (1+C_{base})}$ to train a spatial graph $G$ even for novel categories. To validate the reason, we first assume that an object detector trained on base categories is given. Then, we analyze how novel categories are classified on the object detector as shown in Fig. \ref{fig:bs2nv}. In contrast to the well-classified objects within base categories, we observe that novel objects are classified as similar base categories (e.g., ``cow'' $\rightarrow$ ``dog'', ``horse'' and ``sheep''). This indicates that the representation power for a novel object can be improved by explicitly guiding a network to learn attentive features that are distinguishable from similar base categories. Furthermore, if there are categories that frequently appear together in an image (e.g., ``bicycle'', ``car'' with ``person''), the co-occurrence can also be exploited when learning a novel category (e.g., ``motorbike'' with ``person''). Owing to these advantages, we use a similarity score $c_i$ instead of a raw feature $f_i$ to explore the inter-relationships among objects as expressed in Eq. \ref{eqn:latent_construct}.

\section{Experimental analysis}
\subsection{Benchmark dataset}
We conduct extensive comparisons with existing baselines on few-shot object detection benchmarks \cite{kang2019few,yan2019meta,wang2019meta,wang2020frustratingly,wu2020multi}.
For PASCAL VOC \cite{everingham2010pascal} covering 20 object categories, the novel categories are instance-wise sampled $k$ = 1, 2, 3, 5, 10 times from the trainval of PASCAL VOC 2007+2012 for training. We evaluate on three different novel set split settings: \textbf{Novel set split 1} (\textit{“bird”, “bus”, “cow”, “mbike”, “sofa”/ rest}); \textbf{Novel set split 2} (\textit{“aero”, “bottle”,“cow”,“horse”,“sofa” / rest}) and \textbf{Novel set split 3} (\textit{“boat”, “cat”, “mbike”,“sheep”, “sofa”/ rest}). For MS COCO \cite{lin2014microsoft} covering 80 object categories, the 60 categories disjoint with PASCAL VOC are used as base categories, and the remaining 20 categories are used as novel categories. The novel categories are instance-wise sampled $k$ = 10, 30 times.

\subsection{Mini batch construction}
In the base training phase, there exist large-scale training data for the base categories. If a training image contains novel categories, we regard the region as the background. In the fine-tuning phase, we randomly select instance-wise $k$-shot training examples for both base and novel categories. For a fair comparison, we use identical training examples for the few-shot fine-tuning phase with FSRW \cite{kang2019few}, TFA \cite{wang2020frustratingly} and MPSR \cite{wu2020multi} on both the PASCAL VOC and MS COCO datasets.

\subsection{Implementation details}
We used ResNet-50 \cite{he2016deep} for the feature extractor and RoI Align \cite{he2017mask} for the region proposal.  For all projection functions in the edge regression, we used pre-activated linear layers with the ReLU function. Specifically, we used two graph convolutional layers $L$=2 with the dimensions of 1024 and 256 so that the output dimension $q$ is 256. In addition, we set the scale $\alpha$ to 20 and the category-agnostic bounding box regression layer was adopted. For spatial augmentation, we exploited bi-linear interpolation to resize each region. 

All models were trained using the SGD optimizer with a batch size of 8 on four GPUs, a momentum of 0.9 and a weight decay of 0.001. A learning rate of 0.01 was used during the base training phase. To train the base categories of PASCAL VOC, the model was trained for 240k, 8k, and 4k iterations with learning rates of 0.01, 0.001, and 0.0001, respectively. For MS COCO, the model was trained on base categories for 560k, 140k, and 100k iterations with learning rates of 0.01, 0.001 and 0.0001, respectively. During the few-shot fine-tuning phase on PASCAL VOC with $k \in \{1, 2, 3\}$, we trained the model for 8k and 10k iterations and the learning rates were 0.001 and 0.0001, respectively. For $k \in \{5, 10\}$, we trained the model for 16k and 20k iterations and the learning rates were 0.001 and 0.0001. For MS COCO, we used 100k and 300k iterations for 10- and 30-shots with a learning rate of 0.001.

\begin{table*}[t]
\caption{Few-shot object detection performance (nAP50) of novel categories on the PASCAL VOC 2007 test set. We compare our FSOD-SR with existing methods under the three different novel set splits. nAP50 indicates the mean Average Precision with 0.5 of the IoU threshold and \textbf{BOLD} indicates the state-of-the-art. TFA w/fc and TFA w/cos indicate the TFA \cite{wang2020frustratingly} approach with a fully connected classifier and a cosine classifier, respectively.}\label{Tab:voc_main}
\centering
\begin{adjustbox}{width=1\textwidth, height=2.5cm}
\Huge
\begin{threeparttable}
\begin{tabular}{l|ccccc|ccccc|ccccc}
\hline
                & \multicolumn{5}{c|}{Novel Set Split 1}                                      & \multicolumn{5}{c|}{Novel Set Split 2}                                      & \multicolumn{5}{c}{Novel Set Split 3}                                     \\ \hline
Method/Shot     & 1             & 2             & 3             & 5             & 10            & 1             & 2             & 3             & 5             & 10            & 1             & 2             & 3             & 5             & 10            \\ \hline
FSRW \cite{kang2019few}   & 14.8          & 15.5          & 26.7          & 33.9          & 47.2          & 15.7          & 15.3          & 22.7          & 30.1          & 39.2          & 19.2          & 21.7          & 25.7          & 40.6          & 41.3          \\
MetaDet \cite{wang2019meta}      & 17.1           & 19.1           & 28.9           & 35.0          & 48.8            & 18.2           & 20.6           & 25.9           & 30.6           & 41.5          & 20.1           & 22.3           & 27.9           & 41.9           & 42.9           \\
Meta R-CNN \cite{yan2019meta}        & 19.9          & 25.5          & 35.0            & 45.7          & 51.5          & 10.4          & 19.4          & 29.6          & 34.8          & 45.4 & 14.3          & 18.2          & 27.5          & 41.2          & 48.1         \\
TFA w/fc \cite{wang2020frustratingly}         & 36.8          & 29.1          & 43.6            & 55.7          & 57.0          & 18.2           & 29.0           & 33.4          & 35.5          & 39.0          & 27.7             & 33.6           & 42.5          & 48.7          & 50.2          \\
TFA w/cos \cite{wang2020frustratingly}         & 39.8          & 36.1          & 44.7            & 55.7          & 56.0          & 23.5           & 26.9           & 34.1          & 35.1          & 39.1          & 30.8             & 34.8           & 42.8          & 49.5          & 49.8          \\
MPSR \cite{wu2020multi}   & 41.7          & 42.5          & 51.4          & 55.2          & 61.8          & 24.4          & 29.3          & 39.2          & 39.9          & 47.8          & 35.6          & 41.8          & 42.3          & 48.0          & 49.7          \\

FSOD-SR (ours) & \textbf{50.1} & \textbf{54.4} & \textbf{56.2} & \textbf{60.0} & \textbf{62.4} & \textbf{29.5} & \textbf{39.9} & \textbf{43.5} & \textbf{44.6} & \textbf{48.1}          & \textbf{43.6} & \textbf{46.6} & \textbf{53.4} & \textbf{53.4} & \textbf{59.5} \\ \hline
\end{tabular}
\end{threeparttable}
\end{adjustbox}
\end{table*}

\begin{figure}[t]
\centering
\begin{center}
\includegraphics[width=0.85\linewidth, height=7cm]{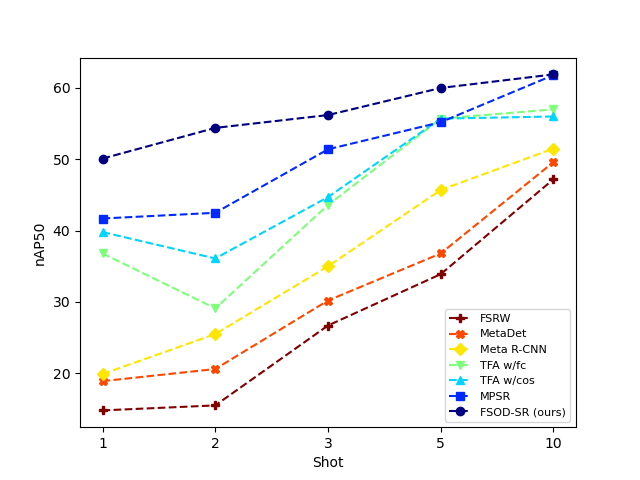}
\end{center}
\caption{nAP50 performance comparison with existing methods using the PASCAL VOC novel set split 1. Existing fine-tuning based models exhibit unstable nAP50 improvements when increasing the number of $k$. The proposed model not only outperforms the previous methods but also achieves stable nAP50 improvements with various $k$-shots.}\label{fig:nAP_plot} 
\end{figure}

\subsection{Results on PASCAL VOC}
We show the nAP50 performance on PASCAL VOC with the three novel set splits in Table \ref{Tab:voc_main}. As shown in the table, our approach outperforms existing methods in all novel set splits and the different numbers of training shots. The improvements become larger as we use the less number of training examples; e.g. 8.4\%, 5.1\% and 8.0\% when compared with MPSR \cite{wu2020multi} on the 1-shot setting of the three novel set splits. To validate this observation clearly, we visualize the nAP50 performance by increasing $k$ and compare it with existing methods. As shown in Fig. \ref{fig:nAP_plot}, the proposed method achieves large performance gains particularly in low-shot settings such as 1-shot and 2-shots. This reveals that when training examples are extremely scarce, the non-local features identified by the spatial reasoning enhance the feature representation of novel RoIs. 

For more detailed comparisons, we provide nAP50, bAP50 and mAP50 in 3-shots and 10-shots with the novel set split 1. After collecting the detection results with all base and novel categories, we separately measured the detection performance on novel categories (nAP), base categories (bAP) and both categories (mAP). As shown in Table \ref{Tab:voc_bsnv}, the proposed method outperforms meta-learning based few-shot objects detectors \cite{kang2019few,yan2019meta} and MPSR \cite{wu2020multi} for the base categories because we freeze the feature extractor during the fine-tuning phase. However, TFA \cite{wang2020frustratingly} exhibits better bAP50 than our method by 1.7\% and 1.0\% in 3-shots and 10-shots, respectively. A possible reason is that novel RoIs affect the feature representation of existing base RoIs during the feature propagation of the GCN.

\begin{table}[t!]
\caption{Generalized object detection performance using the PASCAL VOC novel set split 1. After collecting the detection results of all base and novel categories, we measured nAP50, bAP50 and mAP50 by considering novel categories, base categories and both categories, respectively.}\label{Tab:voc_bsnv}
\centering
\begin{adjustbox}{width=0.7\linewidth}
\tiny
\begin{threeparttable}
\begin{tabular}{c|c|ccc}
\hline
Shot                & Method         & nAP50 & bAP50 & mAP50 \\ \hline
\multirow{5}{*}{3}  & FSRW \cite{kang2019few}  & 26.7  & 64.8  & 55.3  \\
                    & Meta R-CNN \cite{yan2019meta}     & 35.0  & 64.8  & 57.3  \\
                    & TFA w/cos \cite{wang2020frustratingly}      & 44.7  & \textbf{79.1}  & 70.5  \\
                    & MPSR \cite{wu2020multi}           & 51.4  & 67.8  & 63.7  \\
                    & FSOD-SR (ours) & \textbf{56.2}  & 77.4  & \textbf{72.1}  \\ \hline
\multirow{5}{*}{10} & FSRW \cite{kang2019few}  & 47.2  & 63.6  & 59.5  \\
                    & Meta R-CNN \cite{yan2019meta}     & 51.5  & 67.9  & 63.8  \\
                    & TFA w/cos \cite{wang2020frustratingly}      & 56.0  & \textbf{78.4}  & 72.8  \\
                    & MPSR \cite{wu2020multi}           & 61.8  & 71.8  & 69.3  \\
                    & FSOD-SR (ours) & \textbf{62.4}  & 77.4  & \textbf{73.7}  \\ \hline
\end{tabular}
\end{threeparttable}
\end{adjustbox}
\end{table}

\begin{table}[t!]
\begin{center}
\centering
\caption{Few-shot object detection performance on novel categories using the MS COCO 2014 minival set. nAR indicates the mean Average Recall and \textbf{BOLD} means the state-of-the-art. In the case of TFA \cite{wang2020frustratingly}, we used the official pre-trained model to evaluate it under all metrics.}\label{Tab:COCO}
\begin{adjustbox}{width=1\textwidth}
\Huge
\begin{threeparttable}
\begin{tabular}{c|c|c c c|c c c|c c c|c c c}
\hline
Shot       & Method & $nAP$            & $nAP50$          & $nAP75$          & $nAP_S$          & $nAP_M$           & $nAP_L$           & $nAR_1$           & $nAR_{10}$          & $nAR_{100}$         & $nAR_S$          & $nAR_M$           & $nAR_L$           \\ \hline
\multirow{7}{*}{10} & FSRW \cite{kang2019few}            & 5.6           & 12.3          & 4.6           & 0.9          & 3.5           & 10.5          & 10.1          & 14.3          & 14.4          & 1.5          & 8.4           & 28.2          \\ 
                    & MetaDet \cite{wang2019meta}         & 7.1           & 14.6          & 6.1           & 1.0          & 4.1           & 12.2          & 11.9          & 15.1          & 15.5          & 1.7          & 9.7           & 30.1          \\ 
                    & Meta R-CNN \cite{yan2019meta}            & 8.7           & 19.1          & 6.6           & 2.3          & 7.7           & 14.0          & 12.6          & 17.8          & 17.9          & 7.8          & 15.6          & 27.2          \\  
                    & TFA w/fc \cite{wang2020frustratingly}        & 10.0          & 19.2          & 9.2           & 3.9          & 8.4           & 16.3          & 14.8          & 22.2          & 22.4          & 7.8          & 20.3          & 35.0          \\
                    & TFA w/cos \cite{wang2020frustratingly}       & 9.8           & 18.7          & 9.0           & 4.5        & 8.8           & 15.8          & 14.7          & 22.5          & 22.8          & \textbf{9.6}          & 21.1          & 33.9          \\ 
                    & MPSR \cite{wu2020multi}            & 9.8           & 17.9          & 9.7           & 3.3          & 9.2           & 16.1          & 15.7          & 21.2          & 21.2          & 4.6          & 19.6          & 34.3          \\ 
                    & Ours            & \textbf{11.6} & \textbf{21.7} & \textbf{10.4} & \textbf{4.6} & \textbf{10.5} & \textbf{17.2} & \textbf{16.4} & \textbf{23.9} & \textbf{24.1} & {9.3} & \textbf{21.8} & \textbf{37.7} \\ \hline
\multirow{7}{*}{30} & FSRW \cite{kang2019few}           & 9.1           & 19.0          & 7.6           & 0.8          & 4.9           & 16.8          & 13.2          & 17.7          & 17.8          & 1.5          & 10.4           & 33.5          \\ 
                    & MetaDet \cite{wang2019meta}        & 11.3          & 21.7          & 8.1           & 1.1          & 6.2           & 17.3          & 14.5          & 18.9          & 19.2          & 1.8          & 11.1          & 34.4          \\ 
                    & Meta R-CNN \cite{yan2019meta}           & 12.4          & 25.3          & 10.8          & 2.8          & 11.6           & 19.0          & 15.0          & 21.4          & 21.7          & 8.6          & 20.0          & 32.1          \\ 
                    & TFA w/fc \cite{wang2020frustratingly}        & 13.5          & 24.9          & 13.2          & 5.0          & 12.6          & 21.7          & 17.6          & 26.1          & 26.3          & 8.0          & 23.4          & 41.2          \\ 
                    & TFA w/cos \cite{wang2020frustratingly}       & 13.6          & 25.0          & 13.4          & {5.9}          & 12.2          & 21.3          & 17.5          & 26.4          & 26.7          & \textbf{10.1}         & 23.9          & 40.2          \\ 
                    & MPSR \cite{wu2020multi}            & 14.1          & 25.4          & 14.2          & 4.0          & 12.9          & 23.0          & 17.7          & 24.2          & 24.3          & 5.5          & 21.0          & 39.3          \\ 
                    & Ours            & \textbf{15.2} & \textbf{27.5} & \textbf{14.6} & \textbf{6.1} & \textbf{14.5} & \textbf{24.7} & \textbf{18.4} & \textbf{27.1} & \textbf{27.3} & {9.8} & \textbf{25.1} & \textbf{42.6} \\ \hline
\end{tabular}
\end{threeparttable}
\end{adjustbox}
\end{center}
\end{table}

\subsection{Results on MS COCO}
We further evaluate our method on MS COCO with 10-shots and 30-shots. We compare the proposed method with the baselines in multiple metrics as presented in Table \ref{Tab:COCO}. The proposed method consistently outperforms the baselines with different intersection-over-union (IoU) thresholds. Particularly on nAP50, we substantially outperform MPSR \cite{wu2020multi}, observing the gains of 3.8\% and 2.1\% for 10-shots and 30-shots, respectively.

\subsection{Ablation study}

\begin{table}[t!]
\begin{center}
\centering
\caption{Ablation study to clarify the effect of the proposed spatial reasoning and spatial data augmentation. We used the PASCAL VOC novel set split 1. Using the proposed methods together significantly outperforms the plain Faster R-CNN in all cases.}\label{Tab:main_ablation}
\begin{adjustbox}{width=1\textwidth-1mm}
\begin{threeparttable}
\begin{tabular}{c|c|c|c c c c c}
\hline
\multirow{2}{*}{Method}  & \multirow{2}{*}{Spatial Reasoning} & \multirow{2}{*}{Spatial Data Aug} & \multicolumn{5}{c}{Novel Set 1}           \\ \cline{4-8} 
                         &                                    &                                   & 1      & 2      & 3      & 5      & 10     \\ \hline
                     &                                   &                                  & 44.7   & 47.0     & 50.7   & 52.3   & 53.9   \\ 
                     Faster R-CNN
    & \multirow{2}{*}{\checkmark}                 &                 & 47.4   & 48.9   & 53.7   & 58.5   & 59.5   \\ 
                         &                                    &                                   & \textbf{(+2.7)} & \textbf{(+1.9)} & \textbf{(+3.0)} & \textbf{(+6.2)} & \textbf{(+5.6)} \\ \hdashline 
\multirow{2}{*}{FSOD-SR} & \multirow{2}{*}{\checkmark}                 & \multirow{2}{*}{\checkmark}                & 50.1   & 54.4   & 56.2   & 60.0     & 62.4   \\ 
                         &                                    &                                   & \textbf{(+5.4)} & \textbf{(+7.4)} & \textbf{(+5.5)} & \textbf{(+7.7)} & \textbf{(+8.5)} \\ \hline
\end{tabular}
\end{threeparttable}
\end{adjustbox}
\end{center}
\end{table}

\subsubsection{Effect of spatial reasoning and spatial data augmentation}
Table \ref{Tab:main_ablation} shows how each component affects the performance on the PASCAL VOC novel set split 1. As shown in the table, spatial reasoning improves the nAP50 performance for all $k$-shot settings and the performance gain increases as the $k$-shot examples increase. When we further employ the data augmentation strategy, it boosts the gain particularly on extremely low-shot settings (e.g., $k$=1,2) as there are very scarce training examples for each novel category. 

\begin{figure}[t!]
\centering
\begin{center}
\includegraphics[width=0.65\linewidth, height=5.7cm]{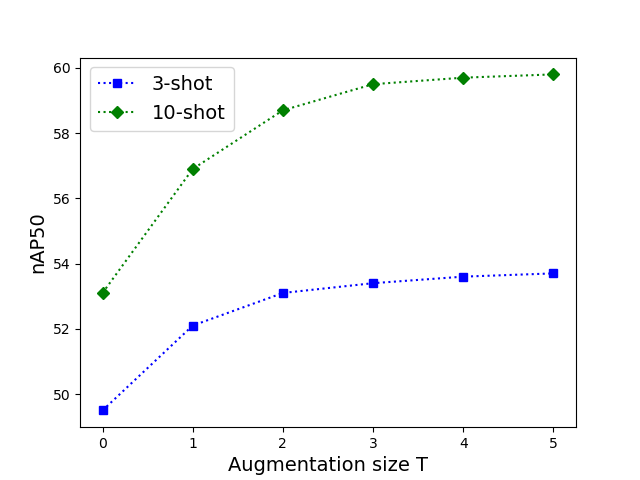}
\end{center}
\caption{Visualization of the nAP50 performance gains with a different $T$ number for spatial data augmentation. $T$ = 0 indicates that data augmentation is not applied. With $T \leq 3$, we can achieve significant performance gains and the gain is saturated at $T > 3$.}\label{fig:aug_ablation}
\end{figure}

\subsubsection{Number of spatial data augmentation}
In the proposed augmentation strategy, we resize each region randomly $T$-times simultaneously. We explore the effect of a different $T$ based on the PASCAL VOC novel set split 3. In Fig. \ref{fig:aug_ablation}, we observe that nAP50 is improved as we increase the number of $T$. However, as the performance gain becomes saturated, we set $T$=3 in all experiments using the augmentation strategy. It is worth noting that as the number of $k$-shot training examples increases (e.g., 3 to 10), the performance gain also increases as more synthetic images are generated.

\begin{table}[t!]
\begin{center}
\centering
\caption{Ablation study on the resizing factor for spatial data augmentation. The result shows that restricting the size of an object so that it does not double its original size is beneficial for object detection.
}\label{Tab:ablation_resizing_factor}
\begin{adjustbox}{width=.55\textwidth}
\begin{threeparttable}
\begin{tabular}{c|ccccc}
\hline
\multirow{2}[4]{*}{Augmentation} & \multicolumn{5}{c}{Novel Set 1} \bigstrut\\
\cline{2-6}  & 1 & 2 & 3 & 5 & 10 \bigstrut\\
\hline
w/o Restriction & 47.7 & 51.3 & 54.7 & 58.9 & 60.3 \bigstrut\\
\hline
w/ Restriction & 50.1 & 54.4 & 56.2 & 60.0 & 62.4 \bigstrut\\
\hline
\end{tabular}%
\end{threeparttable}
\end{adjustbox}
\end{center}
\end{table}

\subsubsection{Resizing factor of spatial data augmentation}
While applying the proposed spatial data augmentation, it is possible for arbitrarily resizing images to make distortion in the spatial context. To prevent from the distortion in our experiments, we restricted the resizing factor to $\sqrt 2$ in width and height. In other words, the area of each region in an image does not become more than twice as large. To validate our approach, we further provide an ablation study on the effect of restricting the object size. As shown in Table \ref{Tab:ablation_resizing_factor}, the performance is clearly degraded if we do not limit the object size for augmentation. We have found that the augmentation without the restriction produces an eccentric shape of an object, such as an extremely fat airplane. We speculate that this causes a distortion to learn contextual relationships or makes an detector to learn out-of-distribution objects.

\begin{table}[t!]
\begin{center}
\centering
\caption{Ablation study on the scaling factor $\alpha$ for the cosine similarity. We  used  the  PASCAL  VOC  novel  set  split  1. The result indicates that there exist an optimal hyper-parameter.}\label{Tab:ablation_sacling_factor}
\begin{adjustbox}{width=.5\textwidth}
\begin{threeparttable}
\begin{tabular}{c|ccc|ccc}
\hline
\multirow{3}[6]{*}{Scale} & \multicolumn{6}{c}{Novel Set 1} \bigstrut\\
\cline{2-7}  & \multicolumn{3}{c|}{bAP50} & \multicolumn{3}{c}{nAP50} \bigstrut\\
\cline{2-7}  & 1 & 3 & 10 & 1 & 3 & 10 \bigstrut\\
\hline
10 & 75.9 & 76.3 & 76.9 & 43.8 & 52.7 & 57.9 \bigstrut\\
\hline
20 & 76.8 & 77.4 & 77.4 & 50.1 & 56.2 & 62.4 \bigstrut\\
\hline
50 & 72.9 & 74.3 & 75.1 & 46.4 & 53.1 & 59.1 \bigstrut\\
\hline
\end{tabular}%
\end{threeparttable}
\end{adjustbox}
\end{center}
\end{table}

\subsubsection{Effect of the scaling factor $\alpha$}
To validate the choice of the scaling factor $\alpha$, we provide the performance using various values to scale up the cosine similarity. Table \ref{Tab:ablation_sacling_factor} shows that $\alpha=20$ consistently outperforms the other scales for all cases. This trend is in line with TFA \cite{wang2020frustratingly} where $\alpha=20$ produces the best performances on novel categories of both datasets.

\subsubsection{Design of edge regression}
As described in Section \ref{sect:SR}, we projected each RoI to an auxiliary classifier to estimate similarity scores with base categories as in Eq. \ref{eqn:latent_construct}. Then, we applied ReLU to create a sparse graph in Eq. \ref{eqn:edge_regression}. To clarify how the design choice affects the performance, we conducted ablation studies on the PASCAL VOC novel set split 2 in Table \ref{Tab:edge_ablation}. ``FC layer'' in the latent embedding column means that we apply a fully connected layer to an RoI feature $f_i$. ``Dense'' in the edge representation column indicates a fully connected edge modeling in which all of the regions are connected. As shown in the table, simply embedding RoI features into a latent space leads to sub-optimal performances as it does not effectively consider the relationships among novel and base categories as discussed in Section \ref{sect:discuss}. In addition, we confirm that the dense edges using redundant connections cause difficulty in training because we have only a few training examples.

\begin{table}[t!]
\begin{center}
\centering
\caption{Ablation study to clarify the design choice for constructing the spatial graph. We used the PASCAL VOC novel set split 2. The fc layer indicates an RoI feature $f_i$ is transformed into a latent space using a fully connected layer and the dense representation indicates that the edges are normalized using the softmax operation instead of the ReLU function.}\label{Tab:edge_ablation}
\begin{adjustbox}{width=0.9\textwidth} 
\begin{threeparttable}
\begin{tabular}{c|c|c|c|ccccc}
\hline
\multicolumn{2}{c|}{Latent Embedding} & \multicolumn{2}{c|}{Edge Representation} & \multicolumn{5}{c}{Novel Set 2} \\ \hline
FC Layer      & Aux Classifier     & Dense            & Sparse           & 1    & 2    & 3    & 5    & 10   \\ \hline
\checkmark             &                     & \checkmark                &                  & 26.9 & 35.8 & 39.9 & 41.2 & 43.5 \\
\checkmark             &                     &                  & \checkmark                & 28.7 & 37.5 & 41.9 & 43.1 & 43.9 \\
              & \checkmark                   & \checkmark                &                  & 27.5 & 36.7 & 41.8 & 42.3 & 44.1 \\
              & \checkmark                   &                  & \checkmark                & \textbf{29.5} & \textbf{39.9} & \textbf{43.5} & \textbf{44.6} & \textbf{48.1} \\ \hline
\end{tabular}
\end{threeparttable}
\end{adjustbox}
\end{center}
\end{table}

\begin{table}[t!]
\begin{center}
\centering
\caption{Ablation study on a cross-dataset generalization ability. The 60 categories of MS COCO were used for training base categories and the 20 categories of PASCAL VOC were fine-tuned for few-shot learning. Then, we evaluated the performance on the PASCAL VOC 2007 test set.}\label{Tab:ablation_coco_to_pascal}
\begin{adjustbox}{width=0.4\linewidth, height=1.9cm}
\begin{tabular}{c|c}
\hline
Method & 20-way 10-shot \bigstrut\\
\hline
FSRW \cite{kang2019few} & 32.3 \bigstrut\\
\hline
MetaDet \cite{wang2019meta} & 34.0 \bigstrut\\
\hline
Meta R-CNN \cite{yan2019meta} & 37.4 \bigstrut\\
\hline
MPSR \cite{wu2020multi} & 42.3 \bigstrut\\
\hline
FSOD-SR (Ours) & 43.2 \bigstrut\\
\hline
\end{tabular}%
\end{adjustbox}
\end{center}
\end{table}

\subsubsection{MS COCO to PASCAL VOC}

Given an object detector trained on MS COCO with 60 base categories using large-scale images, we fine-tuned the detector on 20 novel categories of PASCAL VOC for few-shot learning. Then, we evaluated mAP on the PASCAL VOC 2007 test set. As the base categories of MS COCO are disjoint to the novel categories of PASCAL VOC, this experiment can provide the generalization ability of few-shot learning when the image characteristics are different. As shown in Table \ref{Tab:ablation_coco_to_pascal}, we can observe that the proposed few-shot object detection with spatial reasoning clearly outperforms the recent studies for the cross-dataset generalization problem.

\begin{figure}[t!]
\begin{center}
\includegraphics[width=1\linewidth, height=9cm]{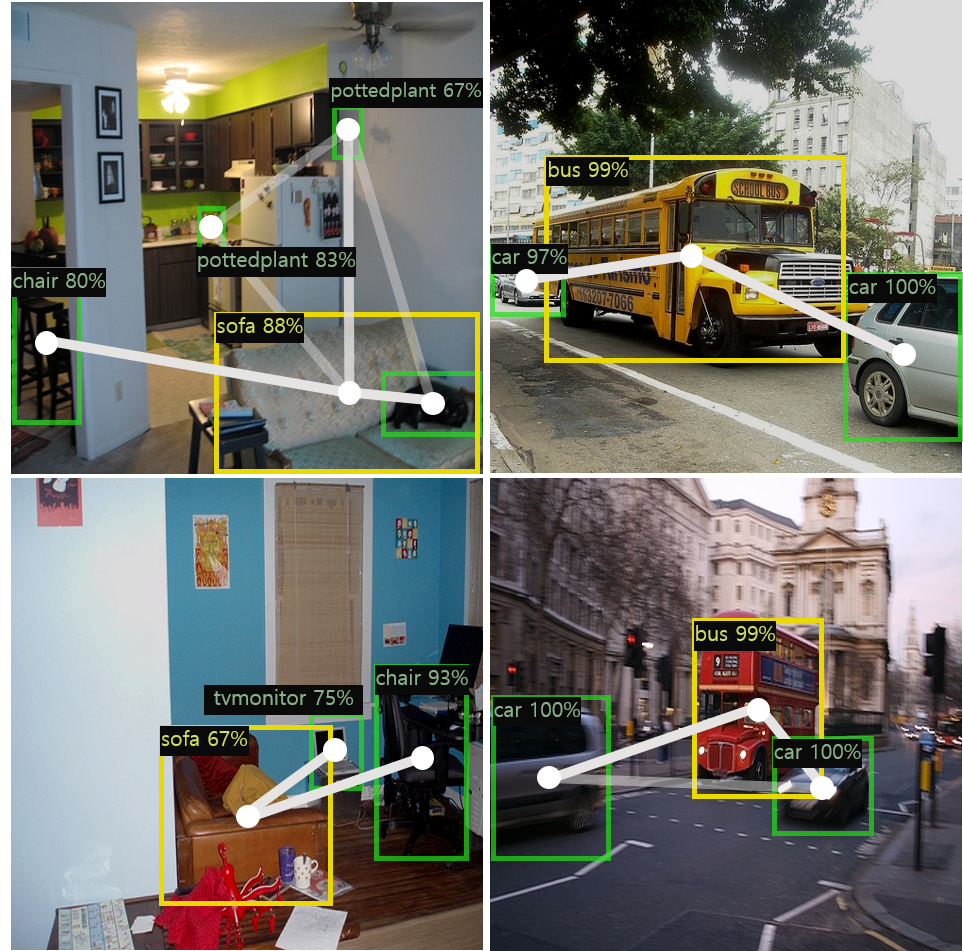}
\end{center}
\caption{Visualization of the spatial graph structure learned by the proposed method. The circle indicates the center of bounding boxes and the line represents the degree of the relatedness between novel and base objects. Novel and base objects are colored yellow and green, respectively. The thicker the line, the more relevant is the connection. Best viewed in color.}
\label{fig:edge_viz}
\end{figure}

\begin{figure}[t!]
\centering
\begin{center}
\includegraphics[width=1.0\linewidth,height=9cm]{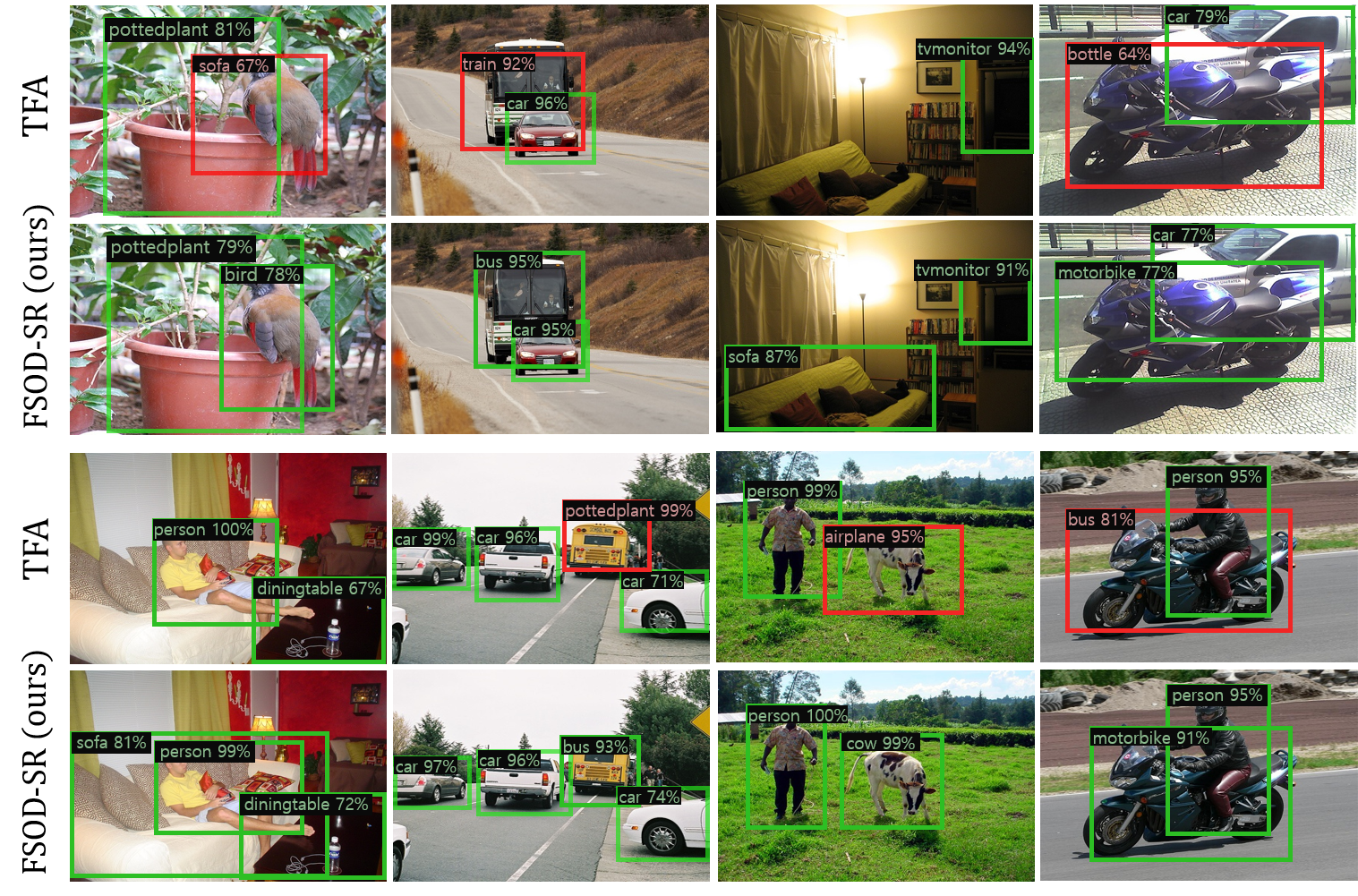}
\end{center}
\caption{Examples of novel objects \textit{(bird, bus, cow, motorbike and sofa)} detected by TFA \cite{wang2020frustratingly} and our FSOD-SR. \textcolor{red}{RED} and \textcolor{green}{GREEN} indicate the failure and success cases, respectively. Our method is more accurate than the existing method because of the spatial reasoning. Best viewed in color.}\label{fig:TF_plot}
\end{figure}

\subsubsection{Visualization}
In Fig. \ref{fig:edge_viz}, we visualize the learned spatial graph structure after the few-shot fine-tuning phase. We can observe that objects with higher probabilities of co-occurrence are well-connected even with novel categories with a few training examples. Furthermore, Fig. \ref{fig:TF_plot} shows that the proposed FSOD-SR detects multiple objects more accurately than TFA \cite{wang2020frustratingly} by considering the context.

\subsection{Limitation}
As shown in Table \ref{Tab:voc_bsnv}, the proposed method degrades the performance on base categories compared to TFA \cite{wang2020frustratingly}. While the feature enhancement by concatenating $e_i$ with a contextual feature $g_i$ and a naive RoI feature $f_i$ significantly boosts the representation power for novel categories, this process inevitably contaminates the feature space for base categories that is well learned on a large-scale dataset. Instead of directly using $e_i$ for classification and box regression, one possible solution is to choose how much we emphasize the newly generated $g_i$ compared to $f_i$ when constructing $e_i$ for Eq. \ref{eqn:cos_cls}. As shown in Fig. \ref{fig:bs2nv}, base categories tend to produce one peaked probability unlike novel categories, and we believe that this different characteristic of classification distribution can be exploited for the purpose. We would like to leave this approach as future work.

\section{Conclusion}
In this paper, we presented a novel few-shot object detection method with spatial reasoning. The goal was to detect novel objects in a context. For the purpose, a spatial graph was defined with the nodes and edges as RoIs and their relatedness relatedness, respectively. The edges were regressed by considering the geometric features of the RoIs and the classification distribution in terms of base categories. Then, we introduced to use a GCN for feature propagation over the constructed spatial graph. To boost the performance further, we proposed spatial data augmentation that exploited the innate property of real-world images. In this approach, each object in an image was resized simultaneously, thereby increasing the number of training images exponentially. Extensive experiments on the widely used PASCAL VOC and MS COCO datasets demonstrated the effectiveness of the proposed FSOD-SR over existing methods. Lastly, we presented a dynamic feature enhancement approach to overcome the performance degradation of base categories that occurred while mixing the original RoI feature and the newly generated contextual RoI feature as future work.

\section*{Acknowledgment}
This work was supported by Institute for Information \& communications Technology Planning \& Evaluation(IITP) grant funded by the Korea government(MSIT) (No. 2017-0-01779, A machine learning and statistical inference framework for explainable artificial intelligence, No. 2019-0-01371, Development of brain-inspired AI with human-like intelligence, and No. 2019-0-00079, Artificial Intelligence Graduate School Program, Korea University).

\bibliography{mybibfile}
\end{document}